\documentclass[conference]{IEEEtran}
\IEEEoverridecommandlockouts

\usepackage{cite}
\usepackage{comment}
\usepackage{amsmath,amssymb,amsfonts}

\usepackage[flushleft]{threeparttable}
\usepackage{algorithmic}
\usepackage{graphicx}
\usepackage{textcomp}
\usepackage{booktabs}
\usepackage{xcolor}
\def\BibTeX{{\rm B\kern-.05em{\sc i\kern-.025em b}\kern-.08em
    T\kern-.1667em\lower.7ex\hbox{E}\kern-.125emX}}
\begin{document}

\title{Machine Unlearning for Speech Question Answering in  Large Audio-Language Models \\
}

\author{\IEEEauthorblockN{Zhe Liu}
\IEEEauthorblockA{\textit{Meta Platforms, Inc.} \\
Menlo Park, USA}
}

\maketitle

\begin{abstract}
Large Audio-Language Models (LALMs) have recently shown strong capabilities in speech understanding and question answering (QA), but they also inherit privacy risks from large-scale training data, including the unintended memorization of sensitive information. In this work, we study machine unlearning for speech QA in LALMs, a setting that is more challenging than prior work on text-based Large Language Models (LLMs) or Automatic Speech Recognition (ASR) due to the tight coupling between acoustic perception and factual knowledge. We present and evaluate multiple unlearning strategies, including gradient ascent, task arithmetic, and alignment-based fine-tuning methods that enforce safe refusal responses, to remove private knowledge while still preserving performance on core capabilities. Through extensive experiments on speech QA datasets, we show that these unlearning methods can reduce the privacy leakage rate by up to 80\% while maintaining near-neutral performance on non-private speech QA and general speech understanding benchmarks. 
\end{abstract}

\begin{IEEEkeywords}
large audio-language models, speech question answering, machine unlearning, privacy preservation.
\end{IEEEkeywords}

\section{Introduction}

Over the past few years, Large Language Models (LLMs) have achieved remarkable success across a wide range of natural language understanding and generation tasks \cite{achiam2023gpt,  grattafiori2024llama, liu2024deepseek}. Motivated by the significant progress in LLMs, the domain of large audio-language models (LALMs) has also undergone a revolutionary transformation and shown promising capabilities for speech and audio understanding and generation \cite{wu2023decoder, xu2025qwen25, xu2025qwen3, defossez2024moshi}. LALMs can be broadly divided into two paradigms: (1) encoder-LLM architectures, where a speech encoder produces continuous representations that are projected into the input space of an LLM backbone; and (2) speech token-based architectures, where audio is discretized into token sequences and modeled directly by the LLM in a unified autoregressive framework.

With the widespread deployment of these models, concerns about privacy risks have also emerged. Studies have demonstrated that private information contained in the training data of LLMs can be extracted by adversaries \cite{carlini2021extracting, huang2022large}. In particular, LLMs have been found to \emph{unintentionally memorize} training examples and, when being appropriately prompted, can emit the memorized content verbatim \cite{carlini2022quantifying}. This is not desirable since it may expose individuals' information. For LALMs, recent investigations also show the unintended memorization of speech training examples for Automatic Speech Recognition (ASR) \cite{liu2025unlearning}. This suggests that LALMs are not exempt from these privacy risks.

Under data protection regulations such as the \emph{Right To Be Forgotten} (RTBF) \cite{mantelero2013eu}, individuals may request the removal of their personal data from deployed models of text-based LLMs or LALMs. While one could simply retrain the models on the remaining data after each deletion request, this becomes prohibitively expensive for large-scale models. The field of \emph{machine unlearning} \cite{bourtoule2021machine, nguyen2025survey, liu2025rethinking} aims to erase targeted information from a trained model without the cost of retraining from scratch. In recent years, post-hoc approximation methods have gained popularity in unlearning text-based LLMs, including \emph{gradient ascent} \cite{jang2023knowledge} and \emph{task arithmetic} \cite{ilharco2022editing}.

While recent efforts have explored unlearning in LLM-based ASR models \cite{liu2025unlearning}, applying unlearning to LALMs for speech question-answering (QA) remains unexplored. Unlearning in speech QA is fundamentally distinct from the ASR setting. ASR is a deterministic sequence-to-sequence mapping task (audio to transcription), where unlearning typically involves suppressing specific vocabulary or phonetic mappings. In contrast, speech QA requires complex factual retrieval conditioned on acoustic input. To successfully unlearn in speech QA, the model must sever the semantic link between an entity and the associated sensitive fact in its parametric memory, without breaking its ability to parse the incoming speech or format a coherent response. Furthermore, as we show empirically, private and non-private questions share largely entangled feature directions in the model's representation space, making surgical erasure of private knowledge particularly challenging.

In this paper, we present and evaluate multiple machine unlearning methods for speech QA in LALMs. We explore the gradient ascent and task arithmetic approaches for this new use case. We additionally investigate alignment-based fine-tuning strategies for unlearning in LALMs that enforce safe refusal behavior, where the model is trained to map private queries to safe refusal responses, effectively overriding memorized answers. In the experiments, we assess success along three axes: the degree to which private information has been erased, the retention of non-private factual knowledge, and the preservation of general-purpose speech and audio understanding capabilities.


To the best of our knowledge, this work is the first systematic study to investigate the unintended memorization problem in LALMs and the corresponding unlearning algorithms for mitigating privacy risks upon data-removal requests. Our contributions are threefold:
\begin{itemize}
    \item We introduce the problem of speech QA unlearning for LALMs, distinguishing it from ASR unlearning by highlighting the entanglement between acoustic perception and factual memorization.
    \item We present multiple unlearning strategies for speech QA, including gradient ascent, task arithmetic, and alignment-based fine-tuning methods that enforce refusal responses, to remove private knowledge while still preserving performance on core capabilities of speech and audio understanding.
    \item We conduct a comprehensive evaluation of these unlearning methods, revealing a fundamental trade-off between private information erasure and non-private knowledge retention. Our results highlight both the effectiveness and inherent limitations of these unlearning techniques in removing deeply memorized knowledge.
\end{itemize}

The rest of the paper is organized as follows. Section~\ref{sec:related} reviews related work. Section~\ref{sec:methods} describes the unlearning methods for speech QA in LALMs. Section~\ref{sec:experiments} presents the experimental setup and results. Finally, Section~\ref{sec:conclusion} concludes the paper.

\section{Related Work}
\label{sec:related}


\subsection{Machine Unlearning in Text-Based Language Models}
Machine unlearning \cite{bourtoule2021machine, nguyen2025survey, liu2025rethinking} seeks to approximate the state of a machine learning model trained without a specific subset of data. In LLMs, exact unlearning (retraining from scratch) is computationally infeasible. Post-hoc approximation methods have thus gained traction. The authors in \cite{jang2023knowledge} present the gradient ascent approach and showed that it is effective in forgetting targeted token sequences in language models. Task Arithmetic \cite{ilharco2022editing} leverages the geometry of the weight space, subtracting task-specific vectors to erase capabilities.

\subsection{Unlearning in Speech Modalities}
The intersection of speech processing and machine unlearning is a nascent field. Prior work by \cite{liu2025unlearning} demonstrates unlearning in the ASR model by erasing specific targeted phrases with the use of gradient ascent, while the authors of \cite{cheng2025speech} conducted experiments on keyword spotting and speaker identification, showing that unlearning speech data is significantly more challenging than unlearning text data. In this work, we study machine unlearning in the speech QA task for LALMs. 

\section{Methods}
\label{sec:methods}

\subsection{Problem Formulation}
Let $\mathcal{M}(\theta_{deployed})$ be an LALM with weights $\theta_{deployed}$, obtained by fine-tuning a base model with weights $\theta_{base}$ on a dataset $\mathcal{D} = \mathcal{D}_{pri} \cup \mathcal{D}_{nonpri}$, where $\mathcal{D}_{pri}$ is the private speech QA data and $\mathcal{D}_{nonpri}$ is the non-private QA data. Assume that each sample of speech QA data takes a combination of audio and text as input and produces a text answer as output.

After the release and deployment of $\mathcal{M}(\theta_{deployed})$, individuals might submit the requests to have their sensitive data $\mathcal{D}_{pri}$ deleted. 

One naive approach for fulfilling such requests is to follow the same training recipe and re-train the model on the corpus of $\mathcal{D}_{nonpri}$. However, this could be computationally costly, especially when the model size is large.

\begin{figure}[htbp]
\centerline{\includegraphics[width=\columnwidth]{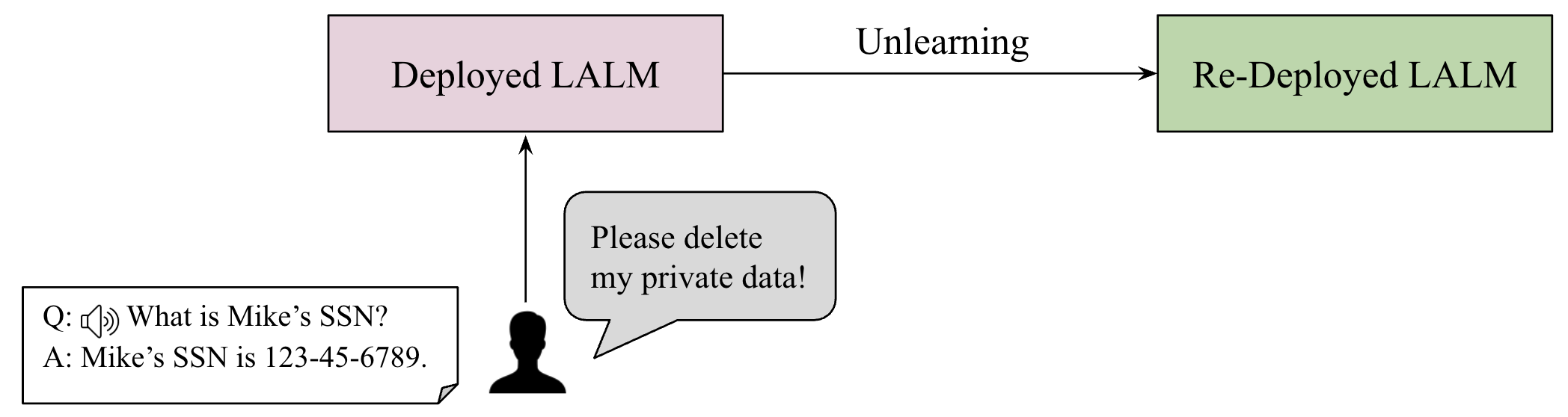}}
\caption{Framework of unlearning LALMs when there are RTBF requests.}
\label{fig:unlearn}
\end{figure}

The objective of machine \emph{unlearning} is to produce optimized weights $\theta_{unlearn}$ such that $\mathcal{M}(\theta_{unlearn})$ exhibits low leakage rate of the private data, while preserving performance in non-private domains and capabilities. Figure~\ref{fig:unlearn} shows the unlearning framework when there are RTBF requests. 

In the following sections, we describe several unlearning approaches applicable to this setting.

\subsection{Gradient Ascent for Unlearning}

Gradient ascent induces forgetting by maximizing the standard auto-regressive language modeling loss on $\mathcal{D}_{pri}$, effectively reversing the gradient direction used during training. This pushes the model's output distribution away from the memorized responses that contain private information.

Specifically, the gradient ascent algorithm is utilized to fine-tune the model $\mathcal{M}(\theta_{deployed})$ on the targeted forget set $\mathcal{D}_{pri}$. The fine-tuning amounts to simply minimizing the likelihood for each text response in the forget set, instead of maximizing it. During the fine-tuning with gradient ascent, we monitor the model's utility as well as the reduction in the level of memorization on the targeted forget set.

\subsection{Task Arithmetic for Unlearning}
Task Arithmetic leverages the observation that fine-tuning a model on a specific task produces a {task vector} in weight space, defined as the difference between the fine-tuned weights and the original pre-trained weights, that encodes the learned capability. To construct the task vector for private information, we fine-tune the base model $\mathcal{M}(\theta_{base})$ exclusively on $\mathcal{D}_{pri}$, yielding weights $\theta_{pri}$. The corresponding task vector is then defined as $\tau_{pri} = \theta_{pri} - \theta_{base}$. Unlearning is performed by subtracting this vector from the model weights $\theta_{deployed}$ (trained on both $\mathcal{D}_{pri}$ and $\mathcal{D}_{nonpri}$), scaled by a factor $\lambda$ that controls the erasure strength:
\begin{equation}
    \theta_{unlearn} = \theta_{deployed} - \lambda\cdot \tau_{pri}
\end{equation}
Intuitively, this negates the direction in weight space associated with privacy memorization. Unlike other unlearning methods, TA is gradient-free at the unlearning stage and the trade-off between forgetting and retention can be tuned post-hoc by adjusting $\lambda$.

\subsection{Safety Supervised Fine-Tuning for Unlearning}
This method performs unlearning by replacing the undesired behavior with a safe alternative. Specifically, we construct a modified version of $\mathcal{D}_{pri}$ in which all privacy-leaking answers are replaced with safe responses that do not disclose sensitive information. The model $\mathcal{M}(\theta_{deployed})$ is then fine-tuned on this modified dataset via standard gradient descent, teaching it to associate private queries with safe responses and overwriting the original question-answer mappings in the model's parameters. For example, a safe response can be a refusal such as ``I cannot help with that request.''

\subsection{Safety Direct Preference Optimization for Unlearning}
Similar to the Safety Supervised Fine-Tuning method above, this method fine-tunes the model $\mathcal{M}(\theta_{deployed})$ on a modified version of $\mathcal{D}_{pri}$, but instead of providing a single target response, it constructs preference pairs. For each audio query, the ``chosen'' response is the safe response (e.g., a refusal), and the ``rejected'' response is the original privacy-leaking answer. The model is then optimized against a frozen reference copy of $\mathcal{M}(\theta_{deployed})$, simultaneously increasing the probability of generating the chosen response and decreasing the probability of the rejected response. The reference model serves as an anchor to prevent excessive drift from the model's general capabilities during optimization.

While gradient ascent and task arithmetic methods aim to remove private information from model parameters, the two alignment-based approaches above operate at the behavioral level by discouraging the model from generating sensitive content. We adopt a practical definition of unlearning focused on preventing privacy leakage at inference time, and therefore include both parametric and behavioral methods under a unified unlearning framework, despite their different mechanisms.

\section{Experiments}
\label{sec:experiments}

\subsection{Experimental Setup}

Our experiments proceed in two phases. In the first phase, we take \texttt{Qwen2.5-Omni-7B} \cite{xu2025qwen25} as the base model and fine-tune it on QA pairs, each consisting of an audio question and a text answer, containing factual information about \emph{fictitious} individuals. The training data here is deliberately composed of two categories: personal \emph{private} information and \emph{non-private} information. After the fine-tuned model is deployed, suppose a data deletion request is received for the individual private information. Then in the second phase, we treat the private training data as the \emph{forget} set and apply unlearning approaches to erase the memorized private information from this deployed model. We evaluate success along three axes: the degree to which private information of individuals has been erased, the retention of non-private factual knowledge about individuals, and the preservation of general-purpose speech QA capabilities.

\textbf{Datasets.} We consider the following datasets as our training and test data in the experiments.

\begin{itemize}
    \item \emph{Pri (Private)} dataset: A targeted forget set comprising speech QA samples that query private information, such as phone numbers, addresses, passwords, and salaries. All names and personal details are fictitious and synthetically generated. Text-based QA pairs were generated initially, and the question text was subsequently converted into audio using a Text-to-Speech (TTS) engine. Disjoint sets of TTS voices were utilized to synthesize 1,000 training and 1,000 test audio samples that share the identical underlying question text. For example, a spoken question  might be ``What is Mike's SSN?'' with the text answer being ``Mike's SSN is 123-45-6789.'' The same question text appears in both training and test sets, but spoken by different TTS voices.

    \item \emph{NonPri (Non-Private)} dataset: A benign, non-private set consisting of speech QA samples that query non-sensitive factual information about individuals, such as hobbies, occupation, personal preferences, and favored items. Following the same synthetic generation protocol as the \emph{Pri} dataset, this set also comprises 1,000 training and 1,000 test audio-text QA pairs that share identical underlying question and response texts. For example, a spoken question might be ``What is Mike's favorite hobby?'' with the answer being ``Mike's favorite hobby is painting.'' Again, this question in the training and test sets is generated using different TTS voices.
    
    \item \emph{Chat (AIR-Bench)} benchmark \cite{yang2024air}: A set of speech and audio understanding questions designed to evaluate the ability of LALMs to comprehend complex audio inputs and follow human instructions. We treat it as a general-purpose benchmark to assess the preservation of conversational capabilities. Unlike the synthetic QA pairs, this dataset features diverse, real-world acoustic conditions and open-ended dialogue, ensuring the unlearning process does not degrade the model's core acoustic perception. This dataset contains 1,193 questions on speech and audio understanding, each paired with a text reference answer. This dataset is used for evaluation purposes only.
\end{itemize}

The details of the synthetic QA pair generation method are described below.
\begin{itemize}
\item \emph{Pri} QA pairs: They are generated by first creating a pool of around 800 fictitious individuals with synthetic names spanning diverse cultural backgrounds. For each individual, one or more QA pairs were created covering 17 categories of private information, including address, email, date of birth, SSN, phone number, employer, bank account, salary, credit card, driver's license, and others. The questions employ around 150 distinct natural-language templates, with multiple phrasings per category (e.g., phone number queries range from ``What is {NAME}'s phone number?'' to ``I'm trying to reach {NAME}. What's the best number?''), ensuring the model cannot simply memorize a single question pattern per category. All answers contain entirely fabricated values. 
\item \emph{NonPri} QA pairs: They reuse the same set of fictitious individuals. Instead of sensitive personal information, each QA pair asks about innocuous personal preferences across 24 lifestyle topics such as hobbies, favorite food, travel, music taste, and hypothetical scenarios. This parallel structure, same individuals but non-sensitive content, enables controlled evaluation of unlearning methods: a model should forget private answers while retaining the ability to answer benign questions about the same people.
\end{itemize}
We note that our use of synthetic \emph{Pri} and \emph{NonPri} data introduces a realism gap compared to real-world speech data. However, it provides fine-grained control over the experimental setup, enabling more precise evaluation of unlearning behavior while avoiding ethical and legal concerns associated with handling real sensitive data.

Notice that the unintended memorization of speech training data can occur in two ways: one involves the memorization of speakers' voices, while the other pertains to the memorization of content and underlying text sequences. In our experimental setup, the training and test sets share the same question texts but use different TTS voices, allowing us to test whether memorized content can be retrieved from the model by an adversary using a previously unseen voice.

\begin{figure}[htbp]
\centerline{\includegraphics[width=0.9\columnwidth]{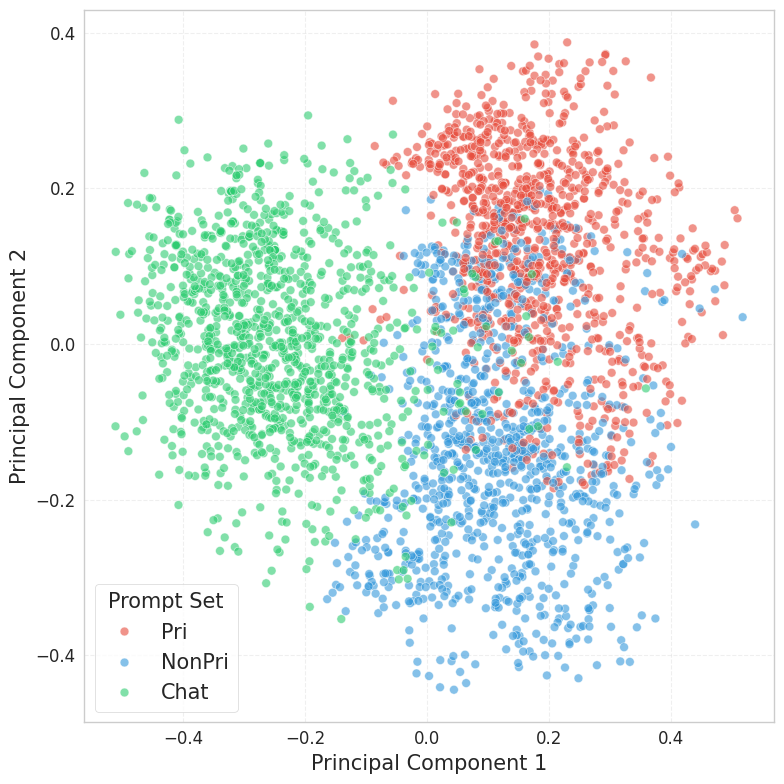}}
\caption{Principal components of representations for the three question sets.}
\label{fig}
\end{figure}

To visualize the structural relationships among the three text question sets above, we extract hidden-state representations from the text-based {Qwen} model and project them onto their first two principal components (Figure~\ref{fig}). The heavy overlap between \emph{Pri} and \emph{NonPri} sets indicates that private and non-private questions share largely entangled feature directions, while the \emph{Chat} cluster is well-separated, as expected given its general-purpose questions.

\textbf{Evaluation Metrics.} For every test sample, we leverage \texttt{Llama3-70B} \cite{grattafiori2024llama} as a judge model to compare LALM's generated response against the reference answer and determine whether the generated response contains the exact same core factual information. We report the percentage of queries that successfully match this criterion. For illustration, we refer to this metric on the \emph{Pri} test dataset as privacy \textit{Leakage Rate} (where lower indicates successful unlearning) and on the \emph{NonPri} or \emph{Chat} test datasets as \textit{Accuracy} (where higher indicates preserved capabilities).

\textbf{Training Strategies.} We train the following components while keeping all other parameters of \texttt{Qwen2.5-Omni-7B} frozen:
\begin{itemize}
    \item For the speech encoder, only the projection adapter is trainable, which maps the speech encodings into token representations compatible with the LLM input space.
    \item For the LLM, the attention layers are trained with Low-Rank Adaptation (LoRA) \cite{hu2022lora} applied to the Q and V projection matrices with rank $r=16$, $\alpha=32$, and a dropout rate of $0.05$.
\end{itemize}
In total, only 12.3M parameters are trainable, accounting for approximately 0.14\% of the total model parameters. All our training utilizes the AdamW optimizer \cite{loshchilov2017decoupled} with a cosine learning rate scheduler.

\textbf{Baselines and Unlearning Approaches.} To recap, in the first phase, we fine-tune the base model \texttt{Qwen2.5-Omni-7B} on the combined \emph{Pri} and \emph{NonPri} training sets; the resulting model is the target for unlearning. In the second phase, we apply unlearning methods to this model. We compare four unlearning strategies against the retraining baseline:
\begin{enumerate}
    \item \textbf{Retrain from Scratch (RT):} The original {Qwen} base model \texttt{Qwen2.5-Omni-7B} is fine-tuned on the \emph{NonPri} training data ($lr=2e^{-4}$), having never been exposed to the \emph{Pri} training data.
    \item \textbf{Gradient Ascent (GA):} The loss function is negated to maximize the loss on the \emph{Pri} training data during the fine-tuning of the model to be unlearned. We use highly constrained learning rates ($1e^{-6}$ and $5e^{-7}$) and gradient clipping to prevent model collapse.
    \item \textbf{Task Arithmetic (TA):} We first fine-tune the original Qwen model \texttt{Qwen2.5-Omni-7B} on the \emph{Pri} training data for the same number of epochs as the model to be unlearned. We then subtract these newly learned weights from the model to be unlearned using a scaling factor $\lambda \in \{0.1, 0.2, 0.3, 0.4, 0.5\}$. This mathematical subtraction is applied simultaneously to the LoRA attention matrices and the audio adapter layer.
    \item \textbf{Safety Supervised Fine-Tuning (SFT):} The model to be unlearned is fine-tuned via standard gradient descent ($lr \in \{1e^{-6}, 5e^{-6}\}$) on the \emph{Pri} training data, but the target response text is replaced with a safe refusal response (e.g., ``I cannot help with that request.'').
    \item \textbf{Safety Direct Preference Optimization (DPO):} We construct preference pairs for the \emph{Pri} training data, where the ``chosen'' response is the safe refusal and the ``rejected'' response is the one with privacy leakage. We set the penalty parameter $\beta=0.1$ and $lr \in \{1e^{-5}, 5e^{-5}\}$. 
\end{enumerate}

\begin{table}[ht!]
  \caption{Evaluation results of the base model and fine-tuned models.}
  \centering
  \resizebox{\columnwidth}{!}{%
  \begin{threeparttable}
  \begin{tabular}{c|l|r|r|r}
    \toprule
    \emph{ID} & \emph{Method} & \emph{Pri} & \emph{NonPri} & \emph{Chat} \\
    \midrule
    B0  & \texttt{BaseModel}  &  0.0\% &  1.7\% & 61.7\% \\
    \midrule
    F5  & \texttt{FT($lr=2e^{-4}$,EP=5)}  &  1.3\% & 28.5\% & 59.3\% \\
    F10  & \texttt{FT($lr=2e^{-4}$,EP=10)} &  5.8\% & 47.8\% & 61.8\% \\
    F20  & \texttt{FT($lr=2e^{-4}$,EP=20)} & 19.5\% & 63.1\% & 61.9\% \\
    \midrule
    F30  & \texttt{FT($lr=2e^{-4}$,EP=30)} & 36.7\% & 63.0\% & 61.5\% \\
    F40  & \texttt{FT($lr=2e^{-4}$,EP=40)} & 49.4\%& 65.3\% & 61.2\% \\
    \bottomrule
  \end{tabular}
  \end{threeparttable}
  }
  \vspace{-0.2cm}
  \label{tab:ft_dynamics}
\end{table}

\subsection{Results}
Table~\ref{tab:ft_dynamics} shows the evaluation results of the base {Qwen} model \texttt{Qwen2.5-Omni-7B} and the fine-tuned models on the combined \emph{Pri} and \emph{NonPri} training datasets with a learning rate of $2e^{-4}$ across varying numbers of epochs.

The base model (B0) has never been exposed to the \emph{Pri} or \emph{NonPri} training data and thus possesses no factual knowledge about these fictitious individuals. As a result, it yields a 0\% privacy leakage rate on the \emph{Pri} test set and a very low 1.7\% accuracy on the \emph{NonPri} test set, although it demonstrates solid speech and audio understanding capabilities on the general \emph{Chat} benchmark (61.7\%).

As fine-tuning progresses, the model rapidly learns the individual factual knowledge on both \emph{Pri} and \emph{NonPri} data. By Epoch 20 (F20), the model reaches a \emph{Pri} leakage rate of 19.5\% alongside a \emph{NonPri} accuracy of 63.1\%. By Epoch 40 (F40), the model obtains 49.4\% \emph{Pri} leakage rate and 65.3\% \emph{NonPri} accuracy. Notably, the general \emph{Chat} capability remains robust throughout training. We select {F20} (moderate leakage rate) and {F40} (high leakage rate) as our candidate models for the subsequent unlearning experiments.

\begin{table}[ht!]
  \caption{Comparison of the fine-tuned models against the retraining baselines.}
  \centering
  \resizebox{\columnwidth}{!}{%
  \begin{threeparttable}
  \begin{tabular}{c|l|r|r|r}
    \toprule
    \emph{ID} & \emph{Method} & \emph{Pri} & \emph{NonPri} & \emph{Chat} \\
    \midrule
    F20  & \texttt{FT($lr=2e^{-4}$,EP=20)} & 19.5\% & 63.1\% & 61.9\% \\
    R20  & \texttt{RT($lr=2e^{-4}$,EP=20)} & 0.3\% & 65.4\% & 59.4\% \\
    \midrule
    F40  & \texttt{FT($lr=2e^{-4}$,EP=40)} & 49.4\%& 65.3\% & 61.2\% \\    
    R40  & \texttt{RT($lr=2e^{-4}$,EP=40)} & 0.7\% & 70.4\% & 58.7\% \\
    \bottomrule
  \end{tabular}
  \end{threeparttable}
  }
  \vspace{-0.2cm}
  \label{tab:retrain_compare}
\end{table}

Table~\ref{tab:retrain_compare} shows the evaluation results of the retraining baseline at 20 and 40 epochs. Both R20 and R40 achieve near-zero leakage on the \emph{Pri} test set while maintaining comparable accuracy to their fine-tuned counterparts (F20 and F40) on the \emph{NonPri} and \emph{Chat} test sets. These retrained models establish a theoretical lower bound on \emph{Pri} leakage rate and an upper bound on \emph{NonPri} accuracy.

\begin{table}[ht!]
  \caption{Unlearning performance applied to the F20 checkpoint (moderate privacy leakage).}
  \centering
  \resizebox{\columnwidth}{!}{%
  \begin{threeparttable}
  \begin{tabular}{c|l|r|r|r}
    \toprule
    \emph{ID} & \emph{Method} & \emph{Pri} & \emph{NonPri} & \emph{Chat} \\
    \midrule
    F20  & \texttt{FT(EP=20)}              & 19.5\% & 63.1\% & 61.9\% \\
    \midrule
    \midrule
    G1.5  & \texttt{GA($lr=1e^{-6}$,EP=5)}  & 16.5\% & 65.9\% & 62.4\% \\
    G1.10  & \texttt{GA($lr=1e^{-6}$,EP=10)} &  5.6\% & 60.9\% & 61.4\% \\
    G1.15  & \texttt{GA($lr=1e^{-6}$,EP=15)} &  0.3\% & 48.6\% & 60.4\% \\
    \midrule
    G2.5  & \texttt{GA($lr=5e^{-7}$,EP=5)}  & 20.0\% & 64.5\% & 62.2\% \\
    G2.10  & \texttt{GA($lr=5e^{-7}$,EP=10)} & 16.5\% & 66.2\% & 62.0\% \\
    G2.15  & \texttt{GA($lr=5e^{-7}$,EP=15)} & 11.3\% & 63.2\% & 61.8\% \\
    G2.20  & \texttt{GA($lr=5e^{-7}$,EP=20)} &  5.4\% & 61.1\% & 61.9\% \\
    G2.25  & \texttt{GA($lr=5e^{-7}$,EP=25)} &  1.9\% & 55.1\% & 60.9\% \\
    G2.30  & \texttt{GA($lr=5e^{-7}$,EP=30)} &  0.7\% & 48.9\% & 60.2\% \\
    \midrule
    \midrule
    T1.1 & \texttt{TA($\lambda=0.1$)} & 16.8\% & 63.2\% & 62.3\% \\
    T1.2 & \texttt{TA($\lambda=0.2$)} & 12.5\% & 59.7\% & 62.4\% \\
    T1.3 & \texttt{TA($\lambda=0.3$)} &  6.8\% & 52.1\% & 62.1\% \\
    T1.4 & \texttt{TA($\lambda=0.4$)} &  1.9\% & 40.3\% & 60.9\% \\
    T1.5 & \texttt{TA($\lambda=0.5$)} &  0.9\% & 28.4\% & 60.7\% \\
    \midrule
    \midrule
    S1.5  & \texttt{SFT($lr=5e^{-6}$,EP=5)}  &  2.9\% & 41.2\% & 63.8\% \\
    \midrule
    S2.5  & \texttt{SFT($lr=1e^{-6}$,EP=5)}  & 19.6\% & 64.5\% & 63.5\% \\
    S2.10  & \texttt{SFT($lr=1e^{-6}$,EP=10)} & 16.8\% & 62.7\% & 64.3\% \\
    S2.15  & \texttt{SFT($lr=1e^{-6}$,EP=15)} & 10.6\% & 56.7\% & 64.5\% \\
    S2.20  & \texttt{SFT($lr=1e^{-6}$,EP=20)} &  5.9\% & 47.6\% & 63.8\% \\
    S2.25  & \texttt{SFT($lr=1e^{-6}$,EP=25)} &  1.5\% & 36.1\% & 63.9\% \\
    \midrule
    \midrule
    D1.5  & \texttt{DPO($lr=5e^{-5}$,EP=5)}  &  8.7\% & 56.8\% & 65.1\% \\
    D1.10 & \texttt{DPO($lr=5e^{-5}$,EP=10)} &  7.1\% & 53.6\% & 64.6\% \\  
    D1.15 & \texttt{DPO($lr=5e^{-5}$,EP=15)} &  4.6\% & 49.2\% & 63.5\% \\
    D1.20 & \texttt{DPO($lr=5e^{-5}$,EP=20)} &  2.2\% & 44.2\% & 63.9\% \\
    \midrule
    D2.5  & \texttt{DPO($lr=1e^{-5}$,EP=5)}  & 16.0\% & 62.5\% & 64.2\% \\
    D2.10 & \texttt{DPO($lr=1e^{-5}$,EP=10)} & 13.2\% & 61.2\% & 63.7\% \\
    D2.15 & \texttt{DPO($lr=1e^{-5}$,EP=15)} & 11.7\% & 60.0\% & 64.8\% \\
    D2.20 & \texttt{DPO($lr=1e^{-5}$,EP=20)} &  9.8\% & 57.5\% & 63.5\% \\
    D2.25 & \texttt{DPO($lr=1e^{-5}$,EP=25)} &  8.5\% & 56.0\% & 63.6\% \\
    D2.30 & \texttt{DPO($lr=1e^{-5}$,EP=30)} &  5.9\% & 52.8\% & 63.8\% \\
    D2.35 & \texttt{DPO($lr=1e^{-5}$,EP=35)} &  3.8\% & 47.9\% & 63.4\% \\
    \bottomrule
  \end{tabular}
  \end{threeparttable}
  }
  \vspace{-0.2cm}
  \label{tab:unlearn_f20}
\end{table}

We then evaluate the four unlearning approaches on the F20 checkpoint (moderate privacy leakage), with results detailed in Table~\ref{tab:unlearn_f20}. We have the following findings:

\begin{itemize}
    \item For GA, as training epochs increase, \emph{Pri} leakage rate decreases steadily. At 15 epochs with a learning rate of $5e^{-7}$ (G2.15), leakage drops to 11.3\% while \emph{NonPri} (63.2\%) and \emph{Chat} (61.8\%) remain comparable to the F20 baseline. At 20 epochs (G2.20), leakage further decreases to 5.4\% with \emph{NonPri} still at 61.1\% accuracy. Pushing further to 30 epochs (G2.30) achieves near-zero leakage (0.7\%), but \emph{NonPri} accuracy degrades substantially to 48.9\%, indicating that prolonged gradient ascent damages the retention of non-private factual knowledge about the fictitious individuals.

    \item For TA, as the scaling factor $\lambda$ increases, \emph{Pri} leakage decreases accordingly. At $\lambda=0.2$ (T1.2), leakage drops to 12.5\% while \emph{NonPri} (59.7\%) and \emph{Chat} (62.4\%) remain close to the F20 baseline. Pushing to $\lambda=0.5$ (T1.5) achieves near-zero leakage (0.9\%), but \emph{NonPri} accuracy collapses to 28.4\%, likely because the shared linguistic structure between \emph{Pri} and \emph{NonPri} causes the task vector to suppress factual QA capabilities about the fictitious individuals.

    \item For safety SFT, as training epochs increase, the model increasingly learns to refuse private queries. At 15 epochs with a learning rate of $1e^{-6}$ (S2.15), leakage drops to 10.6\% while \emph{NonPri} (56.7\%) and \emph{Chat} (64.5\%) remain reasonable. Pushing to 25 epochs (S2.25) achieves 1.5\% leakage, but \emph{NonPri} accuracy collapses to 36.1\%, as the model overgeneralizes the refusal behavior to non-private questions.

    \item For safety DPO, longer training progressively reduces \emph{Pri} leakage. At 15 epochs with a learning rate of $1e^{-5}$ (D2.15), leakage drops to 11.7\% while \emph{NonPri} (60.0\%) and \emph{Chat} (64.8\%) remain stable. At 20 epochs (D2.20), leakage further decreases to 9.8\% with \emph{NonPri} at 57.5\% accuracy. Pushing to 35 epochs (D2.35) achieves 3.8\% leakage with \emph{NonPri} at 47.9\% accuracy.

    \item \textbf{Comparison.} All four methods exhibit the same fundamental trade-off: achieving very low \emph{Pri} leakage rate (e.g., below 5\%) comes at the cost of reduced \emph{NonPri} accuracy. At comparable leakage levels, GA retains the highest \emph{NonPri} accuracy among the four methods, followed by TA and DPO which perform similarly, while SFT suffers the most severe degradation. In particular, GA can reduce the leakage rate by more than 70\% while maintaining near-neutral performance on \emph{NonPri} accuracy. Notably, all methods preserve \emph{Chat} accuracy above 60\%, suggesting that the catastrophic forgetting is localized to the in-domain QA distribution rather than the model's general speech and audio understanding capabilities. However, there is a slight downward trend in \emph{Chat} accuracy with prolonged unlearning for GA and TA in particular, indicating that these methods may begin to affect general capabilities if applied too aggressively. It is also worth noting that the unlearning methods differ in how they handle private queries at inference time. GA and TA still attempt to comply with the query but produce incorrect or fabricated answers, whereas safety SFT and DPO learn to explicitly decline the request with a refusal response.

\end{itemize}

\begin{table}[ht!]
  \caption{Unlearning performance applied to the deeply memorized F40 checkpoint (high privacy leakage).}
  \centering
  \resizebox{\columnwidth}{!}{%
  \begin{threeparttable}
  \begin{tabular}{c|l|r|r|r}
    \toprule
    \emph{ID} & \emph{Method} & \emph{Pri} & \emph{NonPri} & \emph{Chat} \\
    \midrule
    F40  & \texttt{FT($lr=2e^{-4}$,EP=40)} & 49.4\% & 65.3\% & 61.2\% \\
    \midrule
    \midrule
    G3.5  & \texttt{GA($lr=1e^{-6}$,EP=5)}  & 48.1\% & 65.9\% & 61.3\% \\
    G3.10 & \texttt{GA($lr=1e^{-6}$,EP=10)} & 24.5\% & 65.1\% & 60.4\% \\
    G3.15 & \texttt{GA($lr=1e^{-6}$,EP=15)} &  9.7\% & 62.1\% & 60.4\% \\
    G3.20 & \texttt{GA($lr=1e^{-6}$,EP=20)} &  2.4\% & 53.2\% & 58.8\% \\
    G3.25 & \texttt{GA($lr=1e^{-6}$,EP=25)} &  1.0\% & 37.0\% & 57.4\% \\
    \midrule
    \midrule
    T2.1 & \texttt{TA($\lambda=0.1$)} & 43.4\% & 63.3\% & 60.9\% \\
    T2.2 & \texttt{TA($\lambda=0.2$)} & 24.3\% & 59.9\% & 61.4\% \\
    T2.3 & \texttt{TA($\lambda=0.3$)} &  8.5\% & 50.4\% & 61.5\% \\
    T2.4 & \texttt{TA($\lambda=0.4$)} &  3.3\% & 42.2\% & 59.8\% \\
    T2.5 & \texttt{TA($\lambda=0.5$)} &  1.2\% & 27.7\% & 58.0\% \\
    \midrule
    \midrule
    S3.5  & \texttt{SFT($lr=5e^{-6}$,EP=5)}  &  7.9\% & 43.7\% & 63.7\% \\
    \midrule
    \midrule
    D3.5  & \texttt{DPO($lr=5e^{-5}$,EP=5)}  & 22.6\% & 61.2\% & 62.4\% \\
    D3.10 & \texttt{DPO($lr=5e^{-5}$,EP=10)} & 16.7\% & 60.3\% & 63.7\% \\
    D3.15 & \texttt{DPO($lr=5e^{-5}$,EP=15)} & 13.7\% & 60.5\% & 64.3\% \\
    D3.20 & \texttt{DPO($lr=5e^{-5}$,EP=20)} & 12.6\% & 58.5\% & 63.9\% \\
    D3.25 & \texttt{DPO($lr=5e^{-5}$,EP=25)} & 10.4\% & 56.9\% & 64.8\% \\
    D3.30 & \texttt{DPO($lr=5e^{-5}$,EP=30)} &  8.6\% & 53.7\% & 63.6\% \\
    \bottomrule
  \end{tabular}
  \end{threeparttable}
  }
  \vspace{-0.2cm}
  \label{tab:unlearn_f40}
\end{table}

Next, we evaluate the four unlearning approaches on the F40 checkpoint (high privacy leakage), with results presented in Table~\ref{tab:unlearn_f40}. Consistent with the findings on F20, GA again achieves the best trade-off between leakage reduction and knowledge retention. Specifically, G3.10 reduces the leakage rate by roughly half while maintaining neutral performance on \emph{NonPri} accuracy, and G3.15 reduces leakage by around 80\% with only a slight drop in \emph{NonPri} accuracy. Similar to the F20 results, SFT again has the highest performance degradation in \emph{NonPri} accuracy, and a slight decline in \emph{Chat} accuracy is observed with prolonged unlearning for the GA and TA methods.

\section{Conclusion}
\label{sec:conclusion}

In this work, we perform the first investigation of machine unlearning for speech QA in LALMs. We present and evaluate four unlearning strategies: gradient ascent, task arithmetic, safety supervised fine-tuning, and safety direct preference optimization. Through extensive experiments on speech QA datasets, we demonstrate that these unlearning methods can reduce the privacy leakage rate by up to 80\% while maintaining near-neutral performance on non-private speech QA and general speech understanding benchmarks. However, achieving near-zero privacy leakage comes at the cost of degraded non-private QA accuracy.

Future directions include exploring more fine-grained unlearning techniques that can better disentangle private from non-private knowledge, as well as extending the evaluation to other LALM architectures.

\bibliographystyle{IEEEtran}
\bibliography{refs}

@inproceedings{yang2024air,
  title={Air-bench: Benchmarking large audio-language models via generative comprehension},
  author={Yang, Qian and Xu, Jin and Liu, Wenrui and Chu, Yunfei and Jiang, Ziyue and Zhou, Xiaohuan and Leng, Yichong and Lv, Yuanjun and Zhao, Zhou and Zhou, Chang and others},
  booktitle={Proceedings of the Annual Meeting of the Association for Computational Linguistics},
  pages={1979--1998},
  year={2024}
}

@article{xu2025qwen3,
  title={Qwen3-{O}mni technical report},
  author={Xu, Jin and Guo, Zhifang and Hu, Hangrui and Chu, Yunfei and Wang, Xiong and He, Jinzheng and Wang, Yuxuan and Shi, Xian and He, Ting and Zhu, Xinfa and others},
  journal={arXiv preprint arXiv:2509.17765},
  year={2025}
}

@article{grattafiori2024llama,
  title={The {L}lama 3 herd of models},
  author={Grattafiori, Aaron and Dubey, Abhimanyu and Jauhri, Abhinav and Pandey, Abhinav and Kadian, Abhishek and Al-Dahle, Ahmad and Letman, Aiesha and Mathur, Akhil and Schelten, Alan and Vaughan, Alex and others},
  journal={arXiv preprint arXiv:2407.21783},
  year={2024}
}

@article{hu2022lora,
  title={{LoRA}: Low-rank adaptation of large language models},
  author={Hu, Edward J and Shen, Yelong and Wallis, Phillip and Allen-Zhu, Zeyuan and Li, Yuanzhi and Wang, Shean and Wang, Liang and Chen, Weizhu and others},
  journal={Proceedings of International Conference on Learning Representations},
  year={2022}
}

@inproceedings{liu2025unlearning,
  title={Unlearning {LLM}-Based Speech Recognition Models},
  author={Liu, Zhe},
  booktitle={Proceedings of Interspeech},
  pages={3214--3218},
  year={2025}
}

@article{cheng2025speech,
  title={Speech unlearning},
  author={Cheng, Jiali and Amiri, Hadi},
  journal={arXiv preprint arXiv:2506.00848},
  year={2025}
}

@inproceedings{bourtoule2021machine,
  title={Machine unlearning},
  author={Bourtoule, Lucas and Chandrasekaran, Varun and Choquette-Choo, Christopher A and Jia, Hengrui and Travers, Adelin and Zhang, Baiwu and Lie, David and Papernot, Nicolas},
  booktitle={IEEE Symposium on Security and Privacy},
  pages={141--159},
  year={2021},
  organization={IEEE}
}

@article{nguyen2025survey,
  title={A survey of machine unlearning},
  author={Nguyen, Thanh Tam and Huynh, Thanh Trung and Ren, Zhao and Nguyen, Phi Le and Liew, Alan Wee-Chung and Yin, Hongzhi and Nguyen, Quoc Viet Hung},
  journal={ACM Transactions on Intelligent Systems and Technology},
  volume={16},
  number={5},
  pages={1--46},
  year={2025},
  publisher={ACM New York, NY}
}

@article{liu2025rethinking,
  title={Rethinking machine unlearning for large language models},
  author={Liu, Sijia and Yao, Yuanshun and Jia, Jinghan and Casper, Stephen and Baracaldo, Nathalie and Hase, Peter and Yao, Yuguang and Liu, Chris Yuhao and Xu, Xiaojun and Li, Hang and others},
  journal={Nature Machine Intelligence},
  volume={7},
  number={2},
  pages={181--194},
  year={2025},
  publisher={Nature Publishing Group UK London}
}

@inproceedings{jang2023knowledge,
  title={Knowledge unlearning for mitigating privacy risks in language models},
  author={Jang, Joel and Yoon, Dongkeun and Yang, Sohee and Cha, Sungmin and Lee, Moontae and Logeswaran, Lajanugen and Seo, Minjoon},
  booktitle={Proceedings of the 61st Annual Meeting of the Association for Computational Linguistics},
  pages={14389--14408},
  year={2023}
}

@article{ilharco2022editing,
  title={Editing models with task arithmetic},
  author={Ilharco, Gabriel and Ribeiro, Marco Tulio and Wortsman, Mitchell and Gururangan, Suchin and Schmidt, Ludwig and Hajishirzi, Hannaneh and Farhadi, Ali},
  journal={Proceedings of International Conference on Learning Representations},
  year={2023}
}

@article{achiam2023gpt,
  title={{GPT}-4 technical report},
  author={Achiam, Josh and Adler, Steven and Agarwal, Sandhini and Ahmad, Lama and Akkaya, Ilge and Aleman, Florencia Leoni and Almeida, Diogo and Altenschmidt, Janko and Altman, Sam and Anadkat, Shyamal and others},
  journal={arXiv preprint arXiv:2303.08774},
  year={2023}
}

@article{xu2025qwen25,
  title={Qwen2.5-{O}mni Technical Report},
  author={Xu, Jin and Guo, Zhifang and He, Jinzheng and Hu, Hangrui and He, Ting and Bai, Shuai and Chen, Keqin and Wang, Jialin and Fan, Yang and Dang, Kai and others},
  journal={arXiv preprint arXiv:2503.20215},
  year={2025}
}

@inproceedings{wu2023decoder,
  title={On decoder-only architecture for speech-to-text and large language model integration},
  author={Wu, Jian and Gaur, Yashesh and Chen, Zhuo and Zhou, Long and Zhu, Yimeng and Wang, Tianrui and Li, Jinyu and Liu, Shujie and Ren, Bo and Liu, Linquan and others},
  booktitle={IEEE Automatic Speech Recognition and Understanding Workshop},
  pages={1--8},
  year={2023},
  organization={IEEE}
}

@article{liu2024deepseek,
  title={Deepseek-v3 technical report},
  author={Liu, Aixin and Feng, Bei and Xue, Bing and Wang, Bingxuan and Wu, Bochao and Lu, Chengda and Zhao, Chenggang and Deng, Chengqi and Zhang, Chenyu and Ruan, Chong and others},
  journal={arXiv preprint arXiv:2412.19437},
  year={2024}
}

@inproceedings{carlini2021extracting,
  title={Extracting training data from large language models},
  author={Carlini, Nicholas and Tramer, Florian and Wallace, Eric and Jagielski, Matthew and Herbert-Voss, Ariel and Lee, Katherine and Roberts, Adam and Brown, Tom and Song, Dawn and Erlingsson, Ulfar and others},
  booktitle={USENIX Security Symposium},
  year={2021}
}

@article{huang2022large,
  title={Are Large Pre-Trained Language Models Leaking Your Personal Information?},
  author={Huang, Jie and Shao, Hanyin and Chang, Kevin Chen-Chuan},
  journal={arXiv preprint arXiv:2205.12628},
  year={2022}
}

@article{carlini2022quantifying,
  title={Quantifying memorization across neural language models},
  author={Carlini, Nicholas and Ippolito, Daphne and Jagielski, Matthew and Lee, Katherine and Tramer, Florian and Zhang, Chiyuan},
  journal={arXiv preprint arXiv:2202.07646},
  year={2022}
}

@article{mantelero2013eu,
  title={The {EU} Proposal for a General Data Protection Regulation and the roots of the ‘right to be forgotten’},
  author={Mantelero, Alessandro},
  journal={Computer Law \& Security Review},
  volume={29},
  number={3},
  pages={229--235},
  year={2013},
  publisher={Elsevier}
}

@article{defossez2024moshi,
  title={Moshi: a speech-text foundation model for real-time dialogue},
  author={D{\'e}fossez, Alexandre and Mazar{\'e}, Laurent and Orsini, Manu and Royer, Am{\'e}lie and P{\'e}rez, Patrick and J{\'e}gou, Herv{\'e} and Grave, Edouard and Zeghidour, Neil},
  journal={arXiv preprint arXiv:2410.00037},
  year={2024}
}

@article{loshchilov2017decoupled,
  title={Decoupled weight decay regularization},
  author={Loshchilov, Ilya and Hutter, Frank},
  journal={Proceedings of International Conference on Learning Representations},
  year={2019}
}

\end{document}